\documentclass[preprint,review,times,12pt]{elsarticle} 
\biboptions{numbers,sort&compress}

\usepackage[bottom=1.25in]{geometry} 

\usepackage{ltablex}
\usepackage{xltabular}
\keepXColumns
\usepackage{graphicx,amsmath,amssymb,indentfirst,caption,subcaption,booktabs,tabularx,float, comment}
\usepackage{multirow}

\graphicspath{{./figures/}}

\usepackage{url}
\let\href\url

\usepackage{hyperref}

\journal{Solar Energy}

\begin{document}

\begin{frontmatter}

\author[label1]{Ojas Sanghi}
\ead{osanghi@sandia.gov}

\author[label1]{Norman Jost \corref{cor1}}
\ead{nrjost@sandia.gov}

\author[label2]{Benjamin G. Pierce}
\ead{bgp12@case.gov}

\author[label3]{Emma Cooper}
\ead{emma.cooper-2@colorado.edu}

\author[label1]{Isaiah Deane}
\ead{ihdeane@sandia.gov}

\author[label1]{Jennifer L. Braid}
\ead{jlbraid@sandia.gov}

\affiliation[label1]{organization={Sandia National Laboratories},
        addressline={1515 Eubank Blvd SE},
        city={Albuquerque},
        state={New Mexico},
        postcode={87123},
        country={USA}}
\affiliation[label2]{organization={Case Western Reserve University},
        addressline={10900 Euclid Ave},
        city={Cleveland},
        state={Ohio},
        postcode={44106},
        country={USA}}
\affiliation[label3]{organization={University of Colorado, Boulder},
        city={Boulder},
        state={Colorado},
        postcode={80309},
        country={USA}}

\cortext[cor1]{Corresponding author. Email: nrjost@sandia.gov; Phone: +1 (505) 868-5115.}
\title{MultiSolSegment: Multi-Channel Segmentation of Overlapping Features in Electroluminescence Images of Photovoltaic Cells}

\begin{abstract}

Electroluminescence (EL) imaging is widely used to detect defects in photovoltaic (PV) modules, and machine learning methods have been applied to enable large-scale analysis of EL images. However, existing methods cannot assign multiple labels to the same pixel, limiting their ability to capture overlapping degradation features. We present a multi-channel U-Net architecture for pixel-level multi-label segmentation of EL images. The model outputs independent probability maps for cracks, busbars, dark areas, and non-cell regions, enabling accurate co-classification of interacting features such as cracks crossing busbars. The model achieved an accuracy of 98\% and has been shown to generalize to unseen datasets. This framework offers a scalable, extensible tool for automated PV module inspection, improving defect quantification and lifetime prediction in large-scale PV systems.
\end{abstract}

\begin{keyword}
photovoltaic cells \sep electroluminescence imaging \sep multi-channel segmentation \sep U-Net \sep defect detection \sep overlapping feature detection \sep PV cell degradation \sep cell cracking
\end{keyword}

\end{frontmatter}



\section{Introduction}

\subsection{Background}

As photovoltaic (PV) systems become an increasingly prevalent source of electricity generation, ensuring their long-term reliability and performance is critical. This has elevated the importance of operations and maintenance (O\&M) strategies, particularly those aimed at efficiently diagnosing degradation and predicting the lifespan of PV plants. Manual inspection methods, while effective at small scale, are impractical for the growing number and size of utility-scale PV installations. As a result, automated inspection and diagnostic tools are rapidly emerging as a key component of modern PV plant management.

Among these tools, electroluminescence (EL) imaging has become a valuable technique for detecting defects in solar modules, revealing microcracks, busbar darkening, and other degradation. Redondo-Plaza \textit{et al.} review its role within the broader PV inspection landscape alongside I-V curves and photoluminescence imaging, noting that EL imaging is able to detect various failures throughout the lifespan of a photovoltaic module \cite{redondo-plazaInspectionTechniquesPhotovoltaic2024}. Recent advances have made EL imaging increasingly practical for field deployment: Redondo-Plaza \textit{et al.} demonstrates a passive luminescence technique that allows EL and PL images to be acquired under normal operating conditions using the current from the PV string itself, avoiding module disconnection and external power supplies and thus making routine imaging more compatible with real-world O\&M workflows \cite{redondo-plazaPassiveElectroluminescencePhotoluminescence2024}. Furthermore, signal processing innovations like two-dimensional wavelet analysis improve the detectability of electrode-cell interface defects \cite{carbonodelarosaDetectionFailuresElectrodephotovoltaic2024}, and FFT-based modulation suppresses noise in daylight outdoor EL images \cite{redondo-plazaElectroluminescenceImagingBased2025}.

Traditionally, EL imaging has been used qualitatively: experts visually inspect the images to identify the presence of features such as cracks, corrosion, or other signs of degradation \cite{bdourComprehensiveEvaluationTypes2020, kontgesReviewFailuresPhotovoltaic2014c}. However, there has been a recent shift toward automated, quantitative analysis of EL images, driven by the need for scalable, reproducible, and data-driven diagnostics in order to evaluate large datasets of EL images in a timely fashion \cite{delpradosantamariaIndoorDaylightElectroluminescence2025, burattiMachineLearningAdvanced2024}. Accurate interpretation of EL images at scale is crucial for future grids with high percentages of PV generation, as it enables better prediction of system health, targeted maintenance, and improved energy yield forecasting \cite{jahnReviewInfraredElectroluminescence2018}.

These developments underscore that as acquisition and processing of EL images become more field-ready, there is a growing need for robust, scalable segmentation models that can exploit this richer imaging data to quantify complex, overlapping degradation modes at the pixel level

\subsection{Related Work}

Initial efforts in quantifying EL image features focused on image-level classification, where entire modules or cells were labeled according to different magnitudes of defect occurrence.

Some approaches used purely classical image processing methods: Pierce \textit{et al.} employed algorithms such as such as ORB \cite{rubleeORBEfficientAlternative2011} and FAST \cite{rostenFusingPointsLines2005a} to classify EL images into various different categorical regimes \cite{pierceIdentifyingDegradationModes2020}.

Meanwhile, other approaches used purely machine learning methods. Bartler \textit{et al.} adapted a VGG-16 network to classify defects into 4 categories \cite{bartlerAutomatedDetectionSolar2018}. Tang \textit{et al.} also classified into 4 categories using a combination of VGG-16, ResNet50, Inception V3, and MobileNet \cite{tangDeepLearningBased2020b}. Finally, Karimi \textit{et al.} built a Convolutional Neural Network (CNN) to classify busbar darkening into 5 categories \cite{karimiGeneralizedMechanisticPV2020}. 

Many approaches used a mixture of classical and deep machine learning methods. Fada \textit{et al.} used Support Vector Machine (SVM), Random Forest (RF) and Artificial Neural Network (ANN) methods to classify images into three degradation categories \cite{fadaElectroluminescentImageProcessing2017a}. Karimi \textit{et al.} used SVMs and CNNs to classify images into 5 categories \cite{karimiFeatureExtractionSupervised2018a}, and observed that CNNs outperform SVM and RF methods when classifying into 3 categories \cite{karimiAutomatedPipelinePhotovoltaic2019a}. Deitsch \textit{et al.} also found that CNNs, this time a VGG-19 network, outperform SVMs when classifying into 2 categories \cite{deitschAutomaticClassificationDefective2019c}. Demirci \textit{et al.} extracted features with deep neural networks and then classified them with classical methods such as SVM, RF, K-Nearest Neighborhood, Decision Tree, and Naive Bayes, achieving high accuracy when classifying into 4 categories and 2 categories \cite{demirciEfficientDeepFeature2021}. Song \textit{et al.} offer a review of the field and test a variety of methods, including VGG-19, ResNet-50, logistic regression, SVM, and RF to classify images into 4 categories \cite{songComprehensiveCaseStudy2024}. Finally, Chen \textit{et al.} compare RF, ResNet, and YOLO models, finding that ResNet-18 and YOLO perform best when classifying images into 5 categories \cite{chenAutomatedDefectIdentification2022b}.

Other efforts identified areas of defect occurrence alongside severity of defect; Su \textit{et al.} accomplished this by using a combination of a Region Proposal Attention Network (RPAN) and a CNN \cite{suDeepLearningBasedSolarCell2021}.

More recently, segmentation techniques have been applied to EL images to enable pixel-level identification of specific features within a module, making it possible to quantify not just whether a defect exists, but how extensive it is (for instance, the percentage of cell area affected by cracks or the number of cracks in a cell). This in turn enables better modeling of the relationship between physical defects and long-term performance degradation.

Initial steps towards segmentation began with classical image processing approaches, such as image registration and subtraction. Kajari-Schröder \textit{et al.} manually annotated cell degradation to understand their impact \cite{kajari-schrsderCriticalityCracksPV2012}, and subtracted two EL images of a module before and after it underwent a mechanical load test, thereby isolating the cracks that formed \cite{kajari-schroderSpatialOrientationalDistribution2011}.

Spataru \textit{et al.} utilized a different approach, thresholding gray-level EL images to identify regions with solar cell cracks \cite{spataruQuantifyingSolarCell2015}. Rodriguez \textit{et al.} employed a variety of methods, including Gabor filters, Principal Component Analysis, and a Random Forest classifier, to detect defects and isolate them with a laser station \cite{rodriguezAutomaticSolarCell2021}. Whitaker \textit{et al.} employ a series of blurs, filters, and component analyses to extract crack features from EL images \cite{whitakerPVCellCracks2020, whitakerPropertiesPVCell2021},

However, classical approaches are typically non-transferrable to other datasets \cite{rahmanDefectsInspectionPolycrystalline2020}. To improve reliability of EL image segmentation, research has moved to apply deep learning techniques like U-Nets \cite{ronnebergerUNetConvolutionalNetworks2015b}, which have been found to be more effective for pixel classification than classical ML methods like Support Vector Machines due to the implicit spatial coherence of a 2D convolution. Balzategui \textit{et al.} and Pratt \textit{et al.} explored the usage of a U-Net to segment EL images with promising results \cite{balzateguiDefectDetectionPolycrystalline2020} but did not make the model available. Sovetkin \textit{et al.} combined a U-Net model with other encoder-decoder models to segment "droplets" and "shunts" in EL images \cite{sovetkinEncoderDecoderSemantic2021a}. Li \textit{et al.} incorporated attention into a U-Net to detect microcracks in EL images \cite{liSemanticSegmentationMethod2025}, albeit with limited data. Otamendi \textit{et al.} applied a combination of three deep learning approaches to segment defects in EL images \cite{otamendiSegmentationCelllevelAnomalies2021}.

Further efforts for EL image segmentation moved beyond single feature segmentation into multi-classification. At the University of Central Florida (UCF), researchers created a highly accurate U-Net model to segment 5 image features within a single model \cite{fioresiAutomatedDefectDetection2022c}. Hijjawi \textit{et al.} benchmarked multiple models on UCF's EL Dataset, finding that the ConvNeXt-V2 model achieves promising results \cite{hijjawiBenchmarkingStudyInstance2026}. 
At Lawrence Berkeley National Laboratory (LBNL), researchers developed and released \textit{pv-vision} \cite{chenAutomaticCrackSegmentation2023, chenPvvision2022}, another highly accurate U-Net model to segment 2 types of cracks, dark areas, and busbars. \textit{PV-Vision} is the current state-of-the-art being used by researchers and industry. Finally, Pratt et al. \cite{prattDefectDetectionQuantification2021a}, present good results for segmentation of different features, but only the dataset is publicly available \cite{prattBenchmarkDatasetDefect2023a}; the model and the trained weights are not available.

Although performant models like \textit{PV-Vision} were significant advances in EL image feature quantification, their approach assumed that features are mutually exclusive, and therefore lacked the ability to capture the overlap of different degradation features within the same cell. For example, when a crack traverses a busbar in an EL image, the model can only assign one label to the overlapping region, resulting in artificial discontinuities in at least one of the target features. 

This presence of artificial discontinuities in feature detection is important, as it distorts crack length and orientation properties and prevents accurate tracing of crack growth and propagation through aging cycles. Such tasks are important, because with accurate information on crack growth, one can correlate such data with power loss over time and develop an understanding for solar cell degradation and performance over time. In turn, this can inform integration of solar energy into the grid. Thus, it is crucial to be able to detect overlapping degradation features in EL images without introducing artificial discontinuities ("multi-label"), which current state-of-the-art methods cannot do. To be clear, such multi-label assignment would reflect overlapping semantic interpretations in EL images, not overlapping material properties.

The segmentation approaches found during the literature review are summarized in Table \ref{tab:el_approaches}. We separate them into different categories based on the approach used for achieving the segmentation of the images and we point out strengths and weaknesses. Our model is included at the end of the table.
\\
\begin{footnotesize}
\begin{xltabular}{\textwidth}{
|>{\raggedright\arraybackslash}p{2cm}
|>{\raggedright\arraybackslash}X
|>{\raggedright\arraybackslash}X
|>{\raggedright\arraybackslash}X|}
\caption{Summary of EL image classification and segmentation approaches. Most prior methods are difficult to reproduce, do not release pretrained weights, and are often validated only on narrow datasets, limiting transferability. MultiSolSegment addresses these gaps by providing a multi-channel, open-source segmentation framework capable of handling overlapping degradation features.}
\label{tab:el_approaches} \\

\hline
\multicolumn{1}{|c|}{\textbf{Category}} &
\multicolumn{1}{|c|}{\textbf{Approach / Paper}} &
\multicolumn{1}{|c|}{\textbf{Strengths}} &
\multicolumn{1}{|c|}{\textbf{Weaknesses}} \\
\hline

\multicolumn{4}{|c|}{\textbf{(1) Image-level Classification (Entire Module / Cell Classified Into Categories)}} \\
\hline
Classical Feature-Based & Pierce et al. (ORB, FAST) \cite{pierceIdentifyingDegradationModes2020} & 
Fast, low computational cost. Uses interpretable, hand-engineered features. &
Non-transferable; brittle to illumination changes, crystallinity differences, and noise. Cannot localize defects; lacks pixel-level information. Most methods are not packaged for reuse. \\
\hline
ML Classification (CNNs) & Bartler et al. (VGG-16) \cite{bartlerAutomatedDetectionSolar2018}; \newline
Tang et al. (VGG-16, ResNet50, InceptionV3, MobileNet) \cite{tangDeepLearningBased2020b}; \newline Karimi et al. (CNN for busbar darkening) \cite{karimiGeneralizedMechanisticPV2020} &
High accuracy for coarse categories; proven deep CNN architectures; handles visual variation better than handcrafted pipelines. &
Still image-level only (no spatial quantification). Requires large labeled datasets; architecture/weights often not available. Difficult to reproduce; many models are not s or easy to use. \\
\hline
Mixed Classical + DL Feature Pipelines & Fada et al. (SVM, RF, ANN) \cite{fadaElectroluminescentImageProcessing2017a}; \newline Karimi et al. (SVM + CNN) \cite{karimiFeatureExtractionSupervised2018a, karimiAutomatedPipelinePhotovoltaic2019a}; \newline Deitsch et al. (VGG-19 vs. SVM) \cite{deitschAutomaticClassificationDefective2019c}; \newline Demirci et al. (ANN + SVM/RF/KNN/etc.) \cite{demirciEfficientDeepFeature2021}; \newline Song et al. (VGG-19, ResNet-50, SVM, RF) \cite{songComprehensiveCaseStudy2024}; \newline Chen et al. (RF, ResNet-18, YOLO) \cite{chenAutomatedDefectIdentification2022b} &
Flexible pipelines; deep features improve over handcrafted ones; systematically benchmarked across multiple architectures; YOLO-based methods add some spatial awareness. &
Pipelines are complex and not end-to-end. Still primarily category-level, not segmentation. Limited generalization across datasets. Weights and code rarely released, limiting reproducibility. \\

\hline
\multicolumn{4}{|c|}{\textbf{(2) Classical Segmentation (Thresholding, Subtraction, Filters)}} \\
\hline
Image Registration / Subtraction & Kajari-Schröder et al. \cite{kajari-schrsderCriticalityCracksPV2012, kajari-schroderSpatialOrientationalDistribution2011} & 
Provides explicit visualization of crack growth before/after stress; requires no training data. &
Only works with paired load-test images. Cannot generalize to most field images; heavily preprocessing-dependent. Not robust or scalable. \\
\hline
Thresholding + Morphology & Spataru et al. \cite{spataruQuantifyingSolarCell2015} &
Simple, explainable, computationally cheap. &
Extremely sensitive to noise, dark-field variation, and crystallinity differences. Does not generalize. Cannot differentiate crack types or overlapping features. \\
\hline
Classical Filters + RF & Rodriguez et al. (Gabor, PCA, RF) \cite{rodriguezAutomaticSolarCell2021}; \newline Whitaker et al. (filters, blurs, component analysis) \cite{whitakerPVCellCracks2020, whitakerPropertiesPVCell2021} &
Works moderately well for specific defect types; interpretable. &
Does not transfer across datasets; requires manual tuning; cannot model overlapping features; brittle to illumination patterns. Not available as reusable tools. \\

\hline
\multicolumn{4}{|c|}{\textbf{(3) Deep Learning Segmentation (U-Net and encoder-decoder models)}} \\
\hline

Single-Feature Segmentation & Balzategui et al. and Pratt et al. (U-Net) \cite{balzateguiDefectDetectionPolycrystalline2020, prattDefectDetectionQuantification2021a}; \newline Sovetkin et al. \cite{sovetkinEncoderDecoderSemantic2021a}, Li et al. \cite{liSemanticSegmentationMethod2025}, Otamendi et al. \cite{otamendiSegmentationCelllevelAnomalies2021} (U-Net variants) &
First successful DL-based segmentation in EL images; captures spatial coherence; good pixel accuracy. &
Usually trained for one or two feature classes. Not robust to overlapping defects. Many models not open-source; no pretrained weights for transfer learning. \\

\hline

Multi-feature classification & Fioresi et al. (UCF) \cite{fioresiAutomatedDefectDetection2022c}; Hijjawi et al. \cite{hijjawiBenchmarkingStudyInstance2026}; \newline pv-vision (LBNL) \cite{chenAutomaticCrackSegmentation2023, chenPvvision2022}  &
Accurate segmentation of multiple degradation types; strong encoder-decoder architecture. &
Assumes mutually exclusive labels; cannot represent overlapping defects. Weights not publicly available, or available but not straightforward to adapt. \\

\hline
\multicolumn{4}{|c|}{\textbf{(5) MultiSolSegment (Ours)}} \\
\hline

\textbf{Multi-channel Segmentation of Multiple Overlapping Defects} & \textbf{MultiSolSegment (ours)} &
Segments multiple degradation modes simultaneously without forcing mutual exclusivity. Handles crack-busbar overlap; extensible to new features. Provides open-source code, pretrained weights, and easy-to-use pipeline. &
Dataset limited to monocrystalline cells; performance on poly-Si remains an open challenge. Extensibility to new EL datasets requires careful normalization and data setup. \\

\hline
\end{xltabular}
\end{footnotesize}

\subsection{Our Work}

In this work, we address limitations of previous segmentation models by introducing a novel multi-channel U-Net model capable of pixel-level multi-label classification, where each pixel in an input EL image can be simultaneously assigned to multiple image features. Our model enables analysis of interactions between distinct features within the same region of a cell; an important step forward in understanding the compound effects of degradation mechanisms. Multi-label classification with U-Net has been applied to other fields and is especially prevalent in medical imaging, where it has been employed to help diagnose eye disease \cite{fuJointOpticDisc2018} and complex datasets of pediatric cancer \cite{chenInstanceSegmentationDense2022}; Azad \textit{et al.} highlight over a 100 different applications of U-Net in the medical field \cite{azadMedicalImageSegmentation2024}. However, its application to PV defect analysis remains nascent. 

We trained our model from scratch using 585 EL images provided by Case Western Reserve University \cite{whitakerPVCellCracks2020,whitakerPropertiesPVCell2021}, Arizona State University, and LBNL \cite{chenBenchmarkCrackSegmentation2022a}. Images were annotated at the pixel level  across four channels: non-cell, busbar, crack, and dark. Our results demonstrate extremely accurate ($>98$\%) reproduction of manual labels, suggesting this method is well-suited for real-world deployment in PV inspection pipelines.

The rest of the paper is organized as follows. Section \ref{sec:methods} provides details on our model training setup and process. Section \ref{sec:dataset} discusses the composition of our training and testing datasets. Section \ref{sec:results} presents our state-of-the-art results, and Section \ref{sec:discussion} analyzes the broader implications of this work.

Del Prado Santamaría et al., in their AI-Driven Perspectives on EL data \cite{delpradosantamariaIndoorDaylightElectroluminescence2025}, highlight two critical obstacles to wider adoption of EL segmentation models: the scarcity of accessible, ready-to-use architectures and the question of robustness under industrial conditions. To address these points, we have trained our multi-channel model on EL images drawn from multiple institutions and resolutions, and we release the fully documented model and accompanying examples as open-source software, thereby ensuring both reproducibility and straightforward integration into real-world PV inspection pipelines.

The full code for the data preparation and the training is available in the \textit{pvcracks} code repository \cite{jostPvcracks2024}. The repository and the readthedocs \href{https://pvcracks.readthedocs.io/en/latest/}{(pvcracks.readthedocs.io/en/latest/)} also contain example Jupyter notebooks that demonstrate how to use the model. Furthermore, the training and test data \cite{sanghiMultiSolSegmentImagesMasks2025} as well as the final model weights \cite{sanghiMultiSolSegmentTrainedModel} are made publicly accessible.

\section{Methods}
\label{sec:methods}

\subsection{Multi-Channeled Masks and Predictions}

\begin{figure}[H]
    \centering
    \includegraphics[width=1\textwidth]{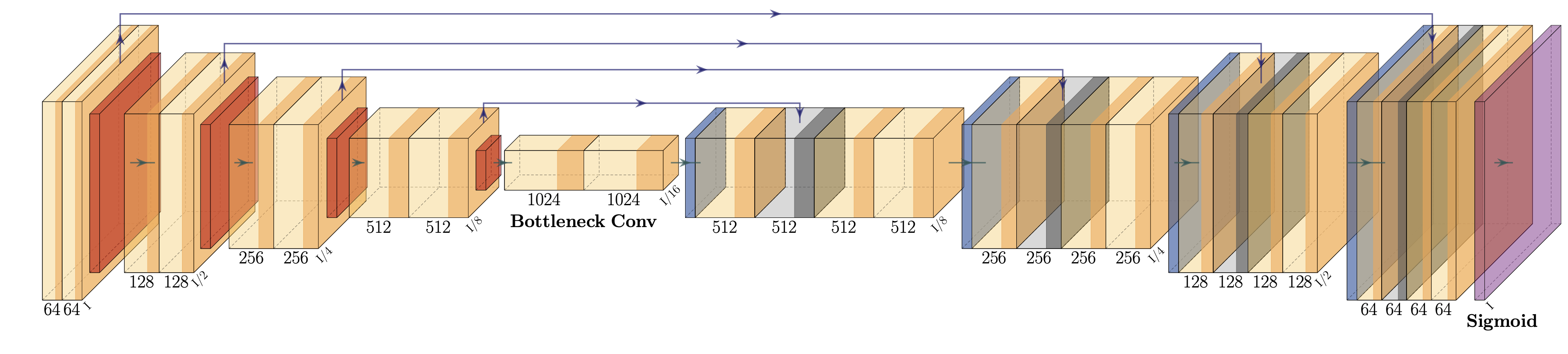}
    \caption{High-level overview of model flow. Made with \cite{iqbalHarisIqbal88PlotNeuralNetV1002018}. Yellow/orange blocks denote convolutional feature maps at each resolution, with block width indicating the number of channels and block height indicating spatial resolution. Darker orange/red inner regions highlight the activated feature representations after convolution and ReLU. Blue-tinted blocks in the decoder indicate upsampled feature maps produced by transposed convolutions. Semi-transparent blocks represent feature concatenation via skip connections between corresponding encoder and decoder stages. The final purple block denotes the output layer with per-channel sigmoid activations, producing independent probability maps for each defect class. Arrows indicate the forward flow of information, while long horizontal connections visualize encoder-decoder skip connections.}
    \label{fig:model_structure}
\end{figure}

The model architecture was a U-Net, following the general encoder-decoder structure described by Ronneberger \textit{et al.} \cite{ronnebergerUNetConvolutionalNetworks2015b}, but constructed to accept 4 input channels. A graphical overview of the model can be seen in Figure \ref{fig:model_structure}. 

The encoder progressively downsamples the input EL image through a sequence of $3\times3$ convolutions (with ReLU activations) and $2\times2$ max-pooling layers, halving the spatial resolution at each stage while doubling the number of feature channels. Starting from a $(H \times W \times 4)$ input tensor, the encoder produces feature maps at progressively smaller resolutions (e.g., $(H/2 \times W/2 \times 64)$, $(H/4 \times W/4 \times 128)$, etc.) until reaching a bottleneck layer with the highest feature dimensionality. The decoder then upsamples these representations using $2\times2$ transposed convolutions, concatenating the result with the corresponding encoder feature maps via skip connections to preserve spatial detail. Convolutions in the decoder progressively reduce the number of channels while restoring the original resolution, culminating in a $(H \times W \times 4)$ output tensor. Furthermore, skip connections between corresponding encoder and decoder layers allow the decoder to access fine-grained details from earlier encoder layers, which helps produce more precise segmentation masks. This helps preserve important spatial information that would otherwise be lost during the encoding process.

A typical U-Net applies a sigmoid on one channel for binary segmentation, or a softmax on multiple channels for multi-class segmentation. In contrast, to enable multi-label segmentation, we applied a per-channel sigmoid activation that converts the model logits into four independent probability maps, one for each defect class. While the core structure remained consistent with a typical U-Net, by adjusting the input and output to apply multiple independent sigmoid activations on multiple channels, we enabled the model to jointly predict pixel-wise classifications for all defect types in a single forward pass. To our knowledge, this is the first application of this approach to EL image classification. 

The U-Net was trained with a 4-channel binary mask as the target and the "Binary Cross Entropy with Logits" loss function. Binary Cross Entropy (BCE) was chosen because it is well-suited to pixel-wise classification tasks like segmentation, where each pixel is treated as a probabilistic prediction. Compared to other loss functions such as Mean Squared Error, BCE produces clearer, more confident segmentation masks and helps the model learn faster and more stably, especially when dealing with imbalanced classes.

The multi channel masks were created using a numpy array of 2D arrays. Each of those 2D arrays contained binary (one-hot) per-pixel activation maps for one of the four classes. The images were labeled using the Supervisely platform. To construct the masks, we followed these steps:

\begin{enumerate}
    \item We downloaded the images and masks (in a \verb|.json| format) from the Supervisely platform.
    \item We converted the \verb|.json| files to \verb|.npy| masks. To accomplish this, we created a multi-hot numpy array; that is, a 3-dimensional numpy array that contained four 2-dimensional one-hot numpy arrays. Each channel corresponds to one class, so in each channel's 2-dimensional array, we marked each pixel with a 1 or 0 depending on whether that pixel was active for that class. This array was saved in a \verb|.npy| file for each image.
    \item After constructing the masks, we split the dataset into training and validation sets in a 4:1 ratio.
    \item After splitting the dataset, we augmented each image and its corresponding mask file by flipping them in the x, y, and both x and y directions. This was conducted only after setting up the splitting the data to prevent any contamination of the test set.
\end{enumerate}

\subsection{Cross-Validation and Hyperparameter Optimization}

To ensure robust model performance and prevent overfitting during hyperparameter selection, we implemented a nested cross-validation strategy using \textit{Ray} for distributed computation and \textit{Ray Tune} for hyperparameter optimization. The outer loop consisted of 5-fold cross-validation to assess model generalization, while the inner loop performed hyperparameter optimization on each training fold. This nested approach provides an unbiased estimate of model performance by ensuring that hyperparameter tuning does not leak information from the test set.

We conducted hyperparameter optimization using a grid search across the following learning rates: \textit{0.0001, 0.0005, 0.001, 0.005, 0.01}. All models were trained for 40 epochs, with early stopping implemented (\textit{patience = 5}) to prevent overfitting.

The final model learning rate, \textit{0.001}, was selected based on the configuration that achieved the best average performance across all five outer folds. Model performance reported in Section \ref{sec:results} corresponds to evaluation of a model trained with this hyperparameter on the final, held-out test set.

\section{Dataset}
\label{sec:dataset}

In this research, four different datasets were used; our data has been made public on the DuraMAT Data Hub \cite{sanghiMultiSolSegmentImagesMasks2025}. All of them contained electroluminescent (EL) images of monocrystalline solar modules, with slight differences in the size and positioning of the solar module, as well as the cropping of the edges. One of the strengths of our ML-based approach to multi-label these images, rather than a classical image processing approach, is that it is agnostic to the formatting specifics of these different types of images. The datasets were provided by researchers from Case Western Reserve University (CWRU), Arizona State University (ASU), and Lawrence Berkeley National Laboratory (LBNL). The CWRU dataset was previously used for crack detection and quantification with classical methods \cite{whitakerPVCellCracks2020,whitakerPropertiesPVCell2021}. The LBNL dataset is part of the \textit{pv-vision} effort \cite{chenOpenDataSets2025}. However, we selected only mono-crystalline cells and had to relabel to train for co-classification. Polycrystalline cells were excluded as they can appear quite different under EL and can have apparent dark spots even in an undegraded state.

To account for intensity and brightness variations across datasets, all images were normalized using a fixed mean and standard deviation ($\mu$, $\sigma$) of (0.485, 0.456, 0.406) and (0.229, 0.224, 0.225), respectively, applied channel-wise. These values are commonly used for RGB images pre-trained on ImageNet and serve to standardize pixel values to a distribution centered near zero with unit variance. Prior to normalization, each image was resized to 256 $\times$ 256 pixels to ensure consistent spatial dimensions. This standardization reduces the impact of dataset-specific brightness or contrast differences, making it less likely that the CNN learns spurious correlations from absolute intensity levels rather than defect morphology.

Each data set was augmented by flipping in the x direction, flipping in the y direction, and flipping in both the x and y directions, effectively increasing the size of our data set by a factor of 4. We then combined the images of all the datasets and randomly split them into training and validation datasets, with 80\% becoming training data and the remaining 20\% becoming validation data. In total, we had 2,340 images, with 1,872 images of training data and 468 images of validation data. This information is broken down in \ref{tab:dataset_info_table}.

\begin{table*}[htp]
    \centering
    \caption{Overview of Datasets Used to Train Model}
    \label{tab:dataset_info_table}
    \begin{tabularx}{\textwidth}{
    | >{\raggedright\arraybackslash}p{4cm}
    | >{\raggedright\arraybackslash}X
    | >{\raggedright\arraybackslash}X
    | >{\centering\arraybackslash}X
    | >{\centering\arraybackslash}X |
    }
        \hline
        \textbf{Dataset Name} &
        \textbf{\# Images} &
        \textbf{\# Total Images (Augmented)} &
        \textbf{\# Images (Training Set)} &
        \textbf{\# Images (Validation Set)} \\
        \hline

        Arizona State University &
        92 & 368 &
        \multirow{5}{*}{\textbf{1,872}} &
        \multirow{5}{*}{\textbf{468}} \\
        \cline{1-3}

        Case Western Reserve University - DuPont &
        123 & 492 & & \\
        \cline{1-3}

        Case Western Reserve University - SunEdison &
        250 & 1000 & & \\
        \cline{1-3}

        Lawrence Berkeley National Laboratory &
        120 & 480 & & \\
        \cline{1-3}

        \textbf{Total:} &
        \textbf{585} &
        \textbf{2340} & & \\
        \hline
    \end{tabularx}
\end{table*}

We had a varying amount of representation for each class in the dataset, as shown in Table \ref{tab:dataset_bias}. Visual examples of these class types can be found in Figure \ref{fig:channeled_high_accuracy}.

\begin{table}[tp]
    \centering
    \caption{Representation of different EL image features in our augmented dataset.}
    \begin{tabularx}{0.7\linewidth}{|X|X|}
        \hline
        \textbf{Class Name} & 
        \textbf{\# Images} \\
        \hline
        Dark        & 364  \\
        Busbar      & 2340  \\
        Crack       & 1688  \\
        Non-Cell    & 2340  \\
        \hline
    \end{tabularx}
    \label{tab:dataset_bias}
\end{table}

Additionally, we calculated numerous statistics about the dataset, detailed in Table \ref{tab:imbalance}:
\begin{table}
    \caption{Dataset characteristics and frequencies for each mask channel}
    \begin{tabularx}{\textwidth}{
        | >{\raggedright\arraybackslash\hyphenpenalty=10000\exhyphenpenalty=10000}p{8cm}
        | >{\raggedright\arraybackslash}X |
    }
        \hline
        \textbf{Property} & \textbf{Value} \\
        \hline
        Cardinality (mean number of labels per pixel)        & 0.241  \\
        \hline
        Density (mean number of labels per pixel / number of labels in the dataset)      & 1.971e-09  \\
        \hline
        Single label pixels (percentage of pixels in the dataset with only one active label)       & 95.073\%  \\
        \hline
        Frequency of each label per image    &  Dark: 15.556\% \newline 
                                                Busbar: 100\% \newline 
                                                Crack: 76.923\% \newline  
                                                Non-cell: 100\%  \\
         \hline
        Frequency of each label per pixel  &    Dark: 1.031\% \newline 
                                                Busbar: 20.259\% \newline 
                                                Crack: 3.569\% \newline  
                                                Non-cell: 7.813\%  \\
         \hline
        Imbalance ratio for each label (\# most frequent label appears / \# label of interest appears) &                       Dark: 0.051 \newline
                                                Busbar: 1 \newline 
                                                Crack: 0.176 \newline  
                                                Non-cell: 0.386  \\
         \hline
        Mean imbalance ratio & 0.403\\
        \hline
    \end{tabularx}
    \label{tab:imbalance}
\end{table}

\section{Results}
\label{sec:results}

\begin{figure}
    \centering
    \includegraphics[width=1\linewidth]{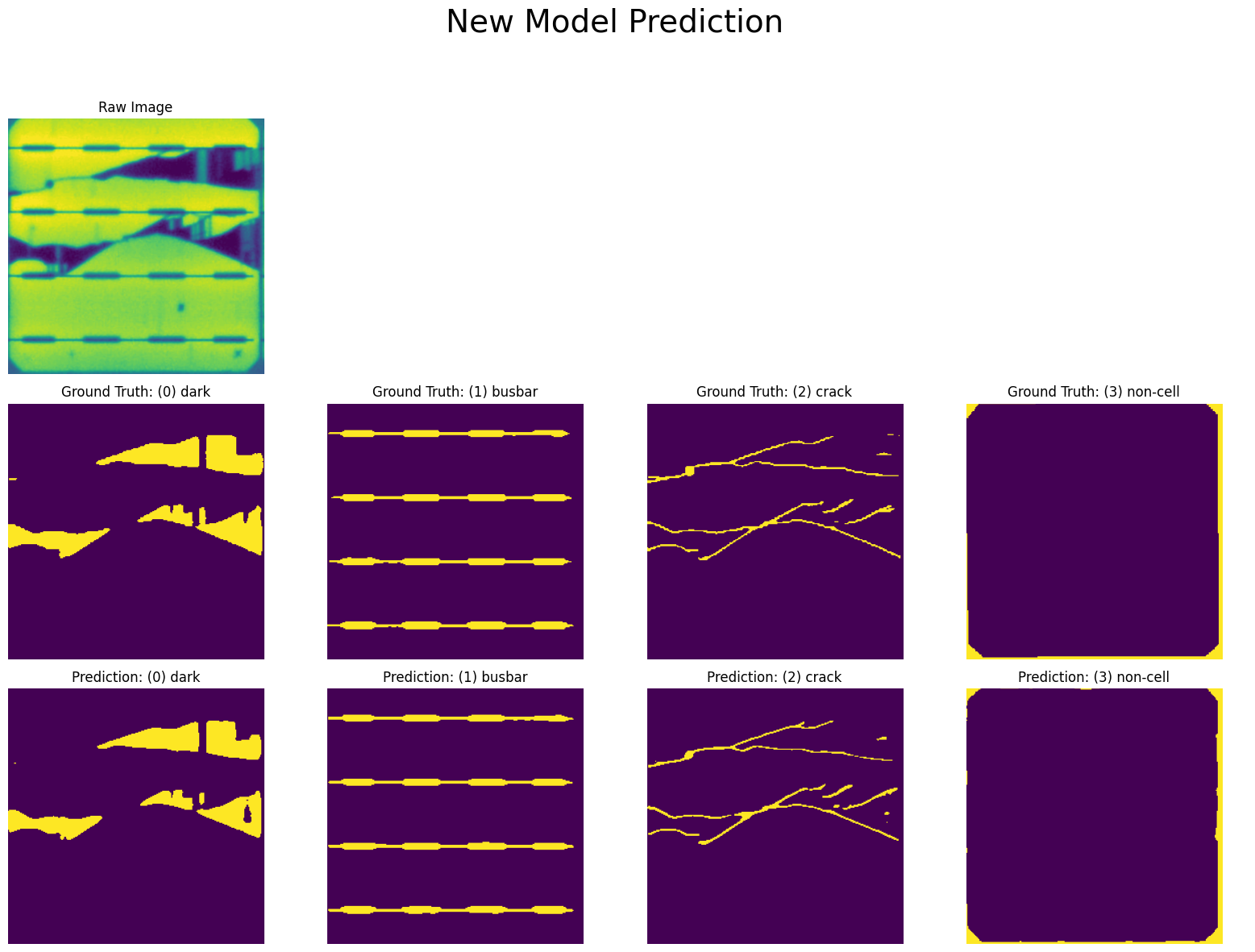}
    \caption{High accuracy results from our model}
    \label{fig:channeled_high_accuracy}
\end{figure}

Segmentation model results were highly accurate, even in complex cases consisting of overlaps between multiple classes, as seen in Figure \ref{fig:channeled_high_accuracy}. For example, the model correctly distinguishes cracks crossing busbars and areas of darkening that partially overlap with non-cell regions. 

Performance was evaluated using multiple metrics: Binary Cross Entropy (BCE) Loss, Accuracy, Precision, Recall, the Dice coefficient, and Intersection over Union. BCE quantifies the per-pixel classification error across all channels, with lower values indicating better performance. Accuracy measures the proportion of correctly classified pixels across the entire image. Precision quantifies the proportion of pixels correctly identified as belonging to the target class among all pixels predicted as that class. Recall measures the proportion of actual target class pixels that were correctly identified by the model. Intersection over Union quantifies the ratio of overlapping area to total area covered by predicted and ground truth (GT) masks. The Dice coefficient, weighs precision and recall equally, making it particularly useful for imbalanced classes such as cracks. 
On our best model, we achieved the scores seen in Table \ref{tab:model_results}.
\newpage 

\begin{table*}[!tp]
    \centering
    \caption{Performance metrics for the best model, averaged across all classes}
    \label{tab:model_results}
    \begin{tabularx}{0.8\linewidth}{|X|X|}
    \hline
         \textbf{Metric}& \textbf{Value} \\
         \hline
         Accuracy (Test Set) & 0.982 \\
         5-Fold CV Accuracy (mean $\pm$ SD) & 0.986740 $\pm$ 0.000461 \\
         Precision & 0.550 \\
         Recall & 0.507 \\
         Dice & 0.503 \\
         Intersection over Union (IoU) & 0.384 \\
    \hline
    \end{tabularx}
\end{table*}

To contextualize our approach, we compare MultiSolSegment against available tools for solar panel defect detection. The primary alternatives for feature classification include PV-Vision \cite{chenAutomaticCrackSegmentation2023}, an open-source tool for automated defect analysis.
A critical distinction of our approach is the simultaneous co-classification of multiple defect types—specifically cracks with busbars and cracks with dark regions—which enables more accurate segmentation in cases where these features overlap or interact spatially.

\begin{figure}[H]
    \centering
    \includegraphics[width=\linewidth]{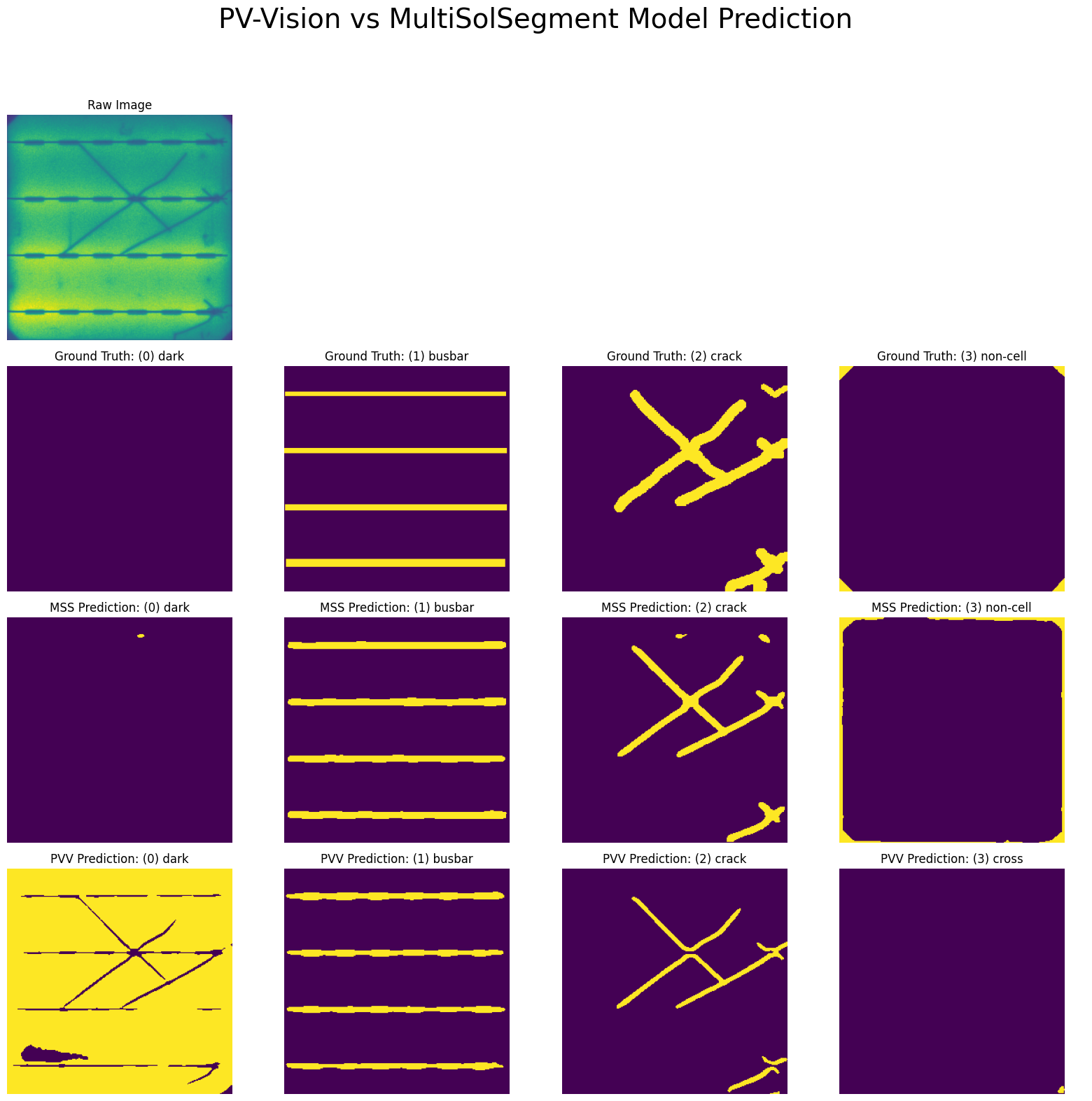}
    \caption{Predictions of MultiSolSegment (MSS) compared to PV-Vision (PVV)}
    \label{fig:pvv_mss_predictions}
\end{figure}

Figure \ref{fig:pvv_mss_predictions} presents a direct comparison between MultiSolSegment and PV-Vision on identical test images. While both models successfully identify busbars and cracks, important differences emerge in their handling of overlapping features. PV-Vision's predictions for the "crack" class introduce artificial discontinuities. At places where the cracks and busbars overlap, PV-Vision mutually exclusively labels only the busbar. In contrast, MultiSolSegment is able to effectly label both classes.

\begin{figure}[H]
    \centering
    \includegraphics[width=\linewidth]{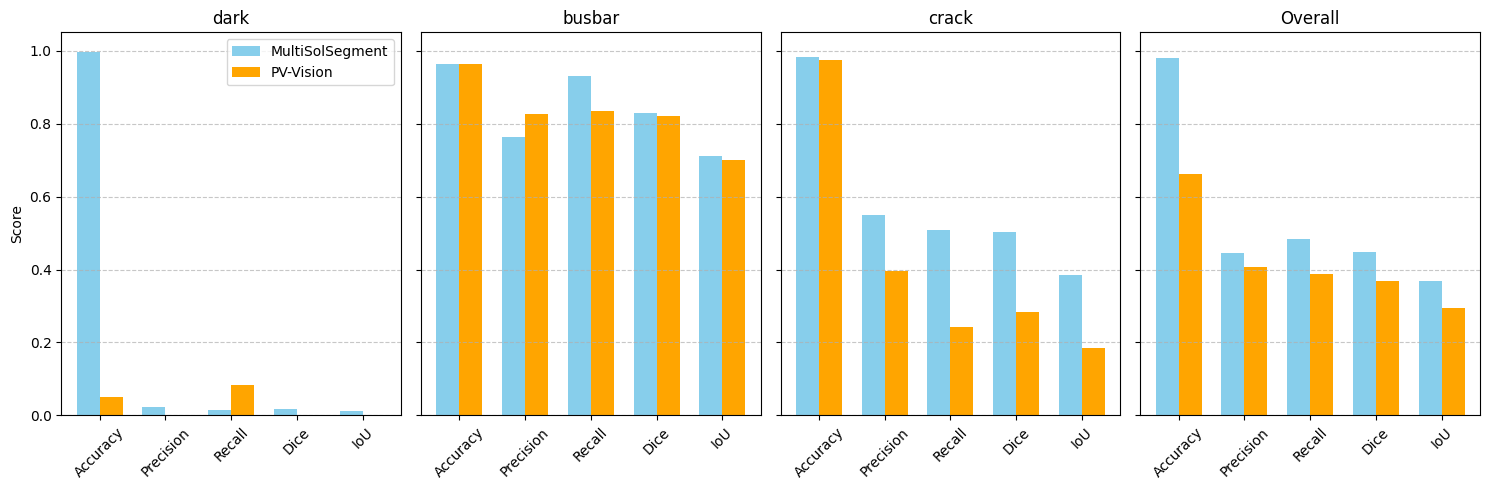}
    \caption{Metrics of MultiSolSegment compared to PV-Vision}
    \label{fig:mss_pvv_metrics}
\end{figure}

The performance comparison across classes (Figure \ref{fig:mss_pvv_metrics}) quantifies these differences. MultiSolSegment outperforms PV-Vision across all classes, in almost every metric.

\begin{figure}[H]
    \centering
    \includegraphics[width=0.75\linewidth]{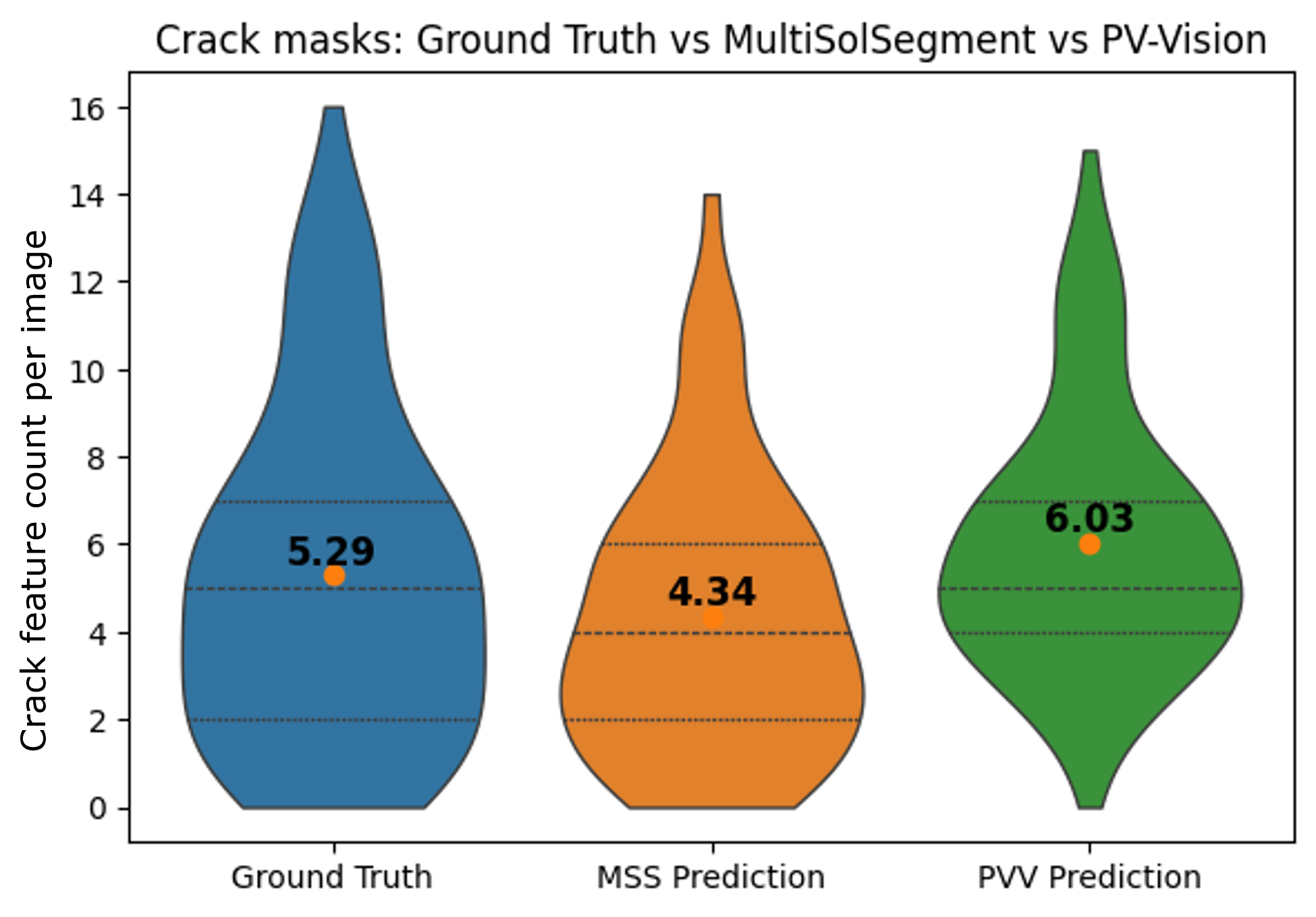}
    \caption{Quantitative detected crack count of MultiSolSegment compared to PV-Vision}
    \label{fig:crack_gt_mss_pvv}
\end{figure}

Finally, Figure \ref{fig:crack_gt_mss_pvv} illustrates the practical impact of these differences. Using the \textit{pvimage} package \cite{karimiAutomatedPipelinePhotovoltaic2019a}, we render violin plots of crack prediction distributions, they include the count of the crack features detected per image and geometrical properties such as perimeter, area, slope... The GT average crack count is at 5.29 per image. MultiSolSegment's count clusters tightly around 4.34, demonstrating consistent slight under-segmentation. In contrast, PV-Vision's predictions spread broadly with a mean count of 6.03, reflecting higher variance and the tendency to fragment crack predictions. This fragmentation occurs because without co-labeling of cracks and busbars, the model incorrectly segments continuous cracks into multiple disconnected components when they intersect structural elements. The tighter distribution of MultiSolSegment's predictions indicates more reliable and consistent crack quantification.

\begin{figure}[H]
    \centering
    \includegraphics[width=\linewidth]{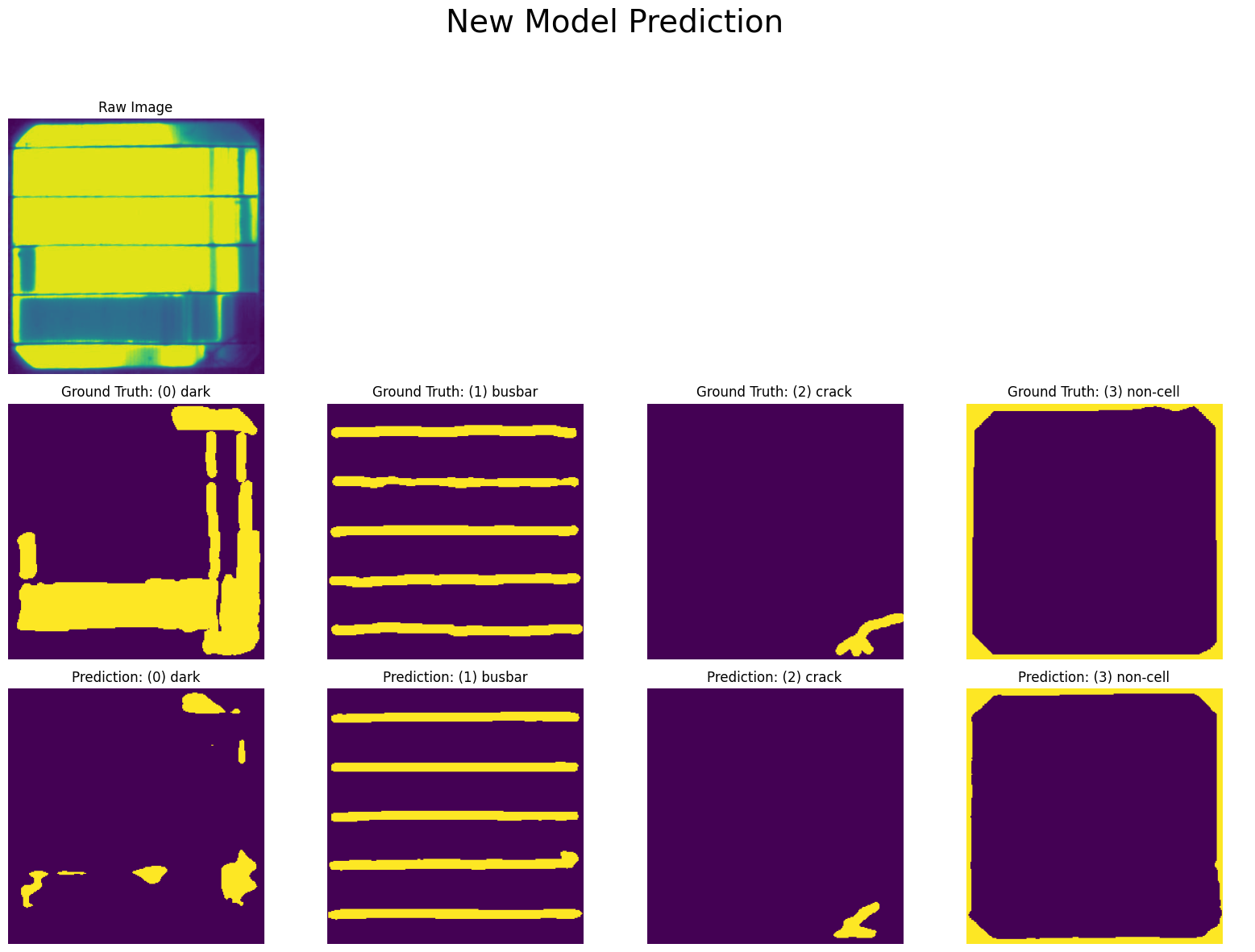}
    \caption{Example of a test case where the model performs poorly}
    \label{fig:channeled_bad_prediction}
\end{figure}

Failure cases were typically confined to specific edge conditions or unusual degradation patterns. For example, darkened regions near image borders were sometimes misclassified as non-cell areas, and in certain heavily degraded modules (Figure \ref{fig:channeled_bad_prediction}), large contiguous dark areas caused the model to under-segment cracks and misclassify portions of the darkened zones. These errors suggest that the model struggles when degradation features overlap strongly or exhibit atypical spatial patterns.

Finally, some areas of low performance, such as in the "dark" class, can be explained by our dataset's class imbalance. For instance, Dark regions represent only 1.031\% of pixels compared to Busbar regions at 20.259\% (Table \ref{tab:imbalance}). However, this imbalance does not compromise the practical utility of our segmentation approach. Dark areas only occur in severely damaged modules, and obtaining real-world samples of EL images with dark spots is rare. Training on this natural distribution ensures our model learns realistic prior probabilities that align with field deployment scenarios. Nevertheless, we acknowledge this as a place of future improvement and discuss it in \ref{sec:future_work}.

\section{Discussion}
\label{sec:discussion}

\subsection{Model Applicability and Extensibility}

Our segmentation model was developed on images from 3 different sources, representing many different types and generations of mono-Si PV cells to capture variations of cell technologies, busbars, and degradation states. These datasets also represent a range of image quality, including both outdoor and laboratory imaging systems and varying image resolution and contrast. Furthermore, we have since applied the segmentation model to a fourth dataset which included additional cell types and busbar quantities not used in training or testing \cite{braidPowerLossEstimation2025}.
The successful application of our segmentation approach to this unseen dataset demonstrates broad extensibility of our model.

In addition to applicability to unseen data, our multi-channel segmentation approach has the additional advantage of being extensible to additional degradation mechanisms and markers in EL images. For example, EL-visible degradation such as overall or gradated darkening associated with potential induced degradation \cite{luoPotentialinducedDegradationPhotovoltaic2017b}, corrosion of busbars and gridlines \cite{fioresiAutomatedDefectDetection2022c, karimiAutomatedPipelinePhotovoltaic2019a}, or oxygen-related ring patterns \cite{chenAutomatedDefectIdentification2022b, vicaristefaniRingDefectsAssociated2025}, can co-occur. Our multi-channel co-classification approach can be expanded to include additional channels for these defects in datasets where they occur. As in this study, training a new model from scratch with additional channels may prove to be the most effective approach for training a model to co-classify new features.

Finally, MultiSolSegment's multi-channel segmentation enables accurate quantification of individual degradation features and their interactions. This per-channel analysis supports power loss prediction from EL images of cracked cells. Previous studies have established correlations between crack detection features and temporal power degradation \cite{jostSolarCellCrack2024a}.

\subsection{Alternative Strategies to Multi-Classification}
\label{sec:alternative_strategies}

In applying the original \textit{pv-vision} model to our data, we found that it performed poorly on cells with $<$5 busbars and modern, thin buswires, as its original training data comprised older, fielded PV modules with fewer, thicker busbars. The modern, thin buswires were misclassified as cracks, and the model puzzlingly misclassified cell corners as cracks as well, despite the original training data also containing this feature. This may have been because the model was mistakenly interpreting the sharp edge in these corners as indicative of a crack. We began our efforts to improve \textit{pv-vision} performance for our dataset with transfer learning. 

\begin{figure}
    \centering
    \includegraphics[width=1\linewidth]{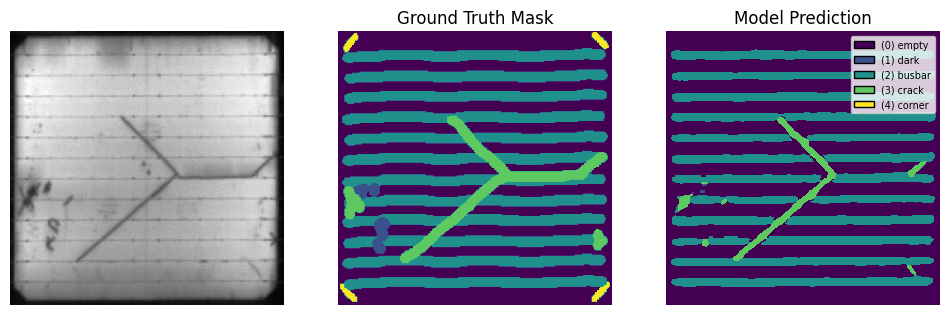}
    \caption{Low accuracy of a transfer-learned pv-vision}
    \label{fig:pv-vision_original_low_accuracy}
\end{figure}

For transfer learning, we first tried to add an additional label for cell corners, first as lines as in Figure \ref{fig:pv-vision_original_low_accuracy} and later as the corner area; the hope was to preserve the model's knowledge for the other classes, while teaching it to not confuse the corners of the images for cracks.  However, the accuracy of this approach was extremely low, as seen in Figure \ref{fig:pv-vision_original_low_accuracy}. 

\begin{figure}
    \centering
    \includegraphics[width=1\linewidth]{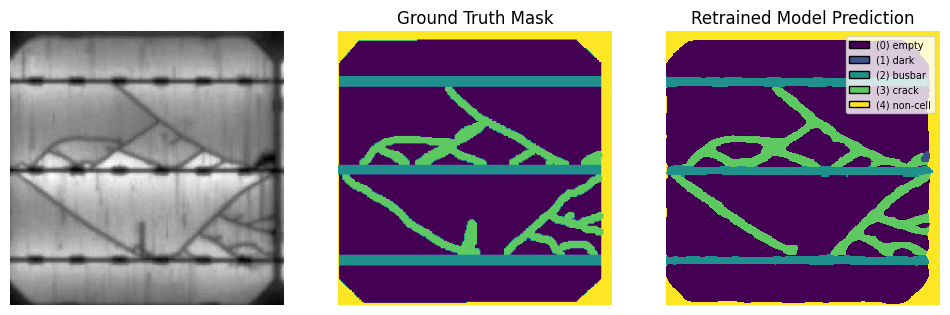}
    \caption{High accuracy of a transfer-learned pv-vision model with the non-cell class}
    \label{fig:pv-vision_non-cell_high_accuracy}
\end{figure}

After realizing that cell cropping often results in non-cell area at the cell edges and not just the corner, we relabeled the crack class to include these edges as a 'non-cell' class. This led to an immediate and stark improvement in accuracy, even in complex cases, as seen in Figure \ref{fig:pv-vision_non-cell_high_accuracy}.

\begin{figure}
    \centering
    \includegraphics[width=0.8\linewidth]{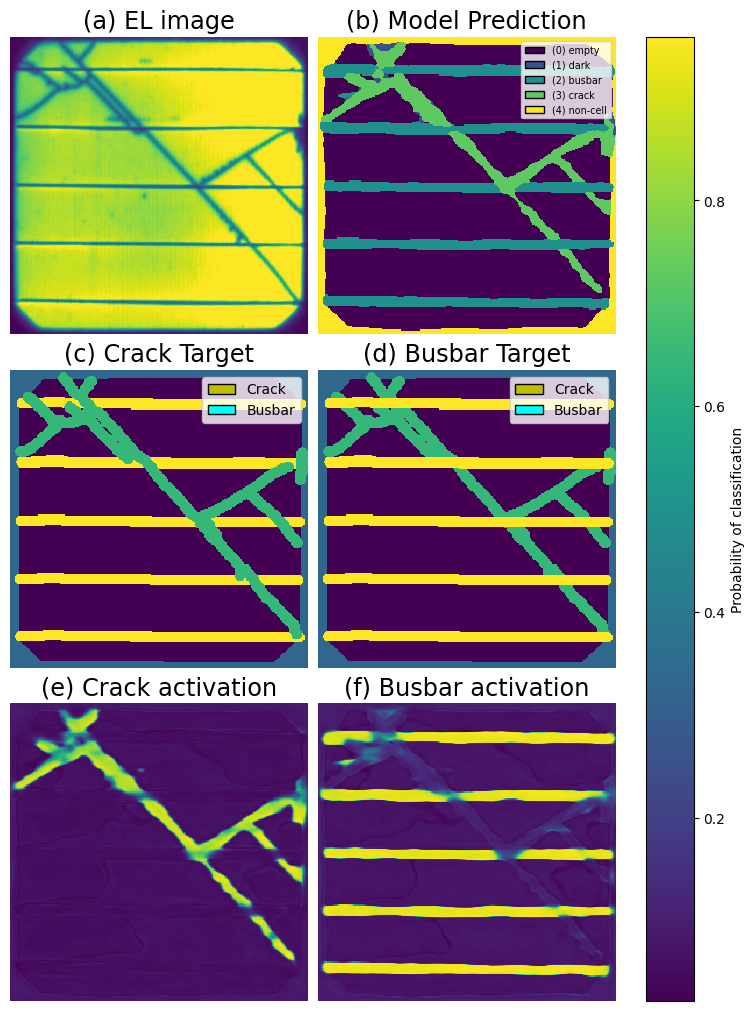}
    \caption{Lack of multi-labeling in our single-channel model. (a) Original EL image. (b) Model prediction using single-channel classification; each pixel assigned to only one of five mutually exclusive classes. (c) Crack target mask from the "double mask" training setup, where overlap pixels are labeled as both crack and busbar. (d) Busbar target mask from the "double mask" training setup. (e) Model activiation map for the crack class, showing predicted probability at each pixel. (f) Model activation map for the busbar class.}
    \label{fig:pv-vision_no_multi_label}
\end{figure}

\begin{figure}
    \centering
    \includegraphics[width=1\linewidth]{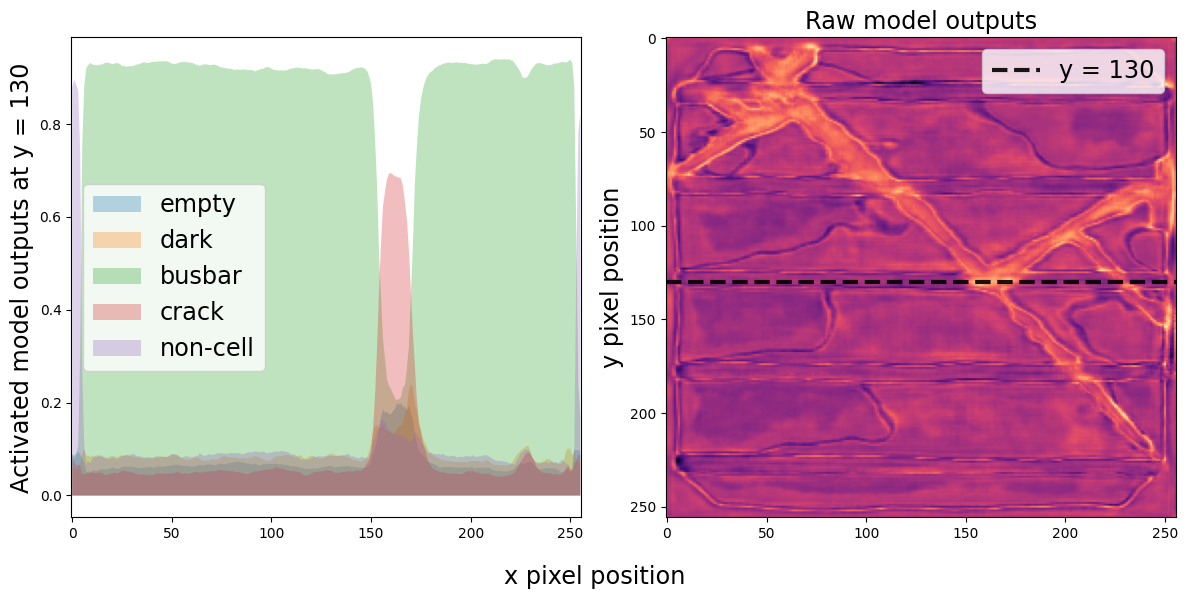}
    \caption{Activation values of pixels at y = 130 for the different classes. On the right is the raw model output for the entire image, with the dashed line indicating y = 130. On the left are the class activation values along this horizontal line.}
    \label{fig:pv-vision_no_multi_label_thresholding}
\end{figure}

Though the single-channel model achieved high overall accuracy, it was fundamentally constrained to predicting a single class per pixel. This limitation became most apparent when cracks overlapped with busbars: rather than labeling such pixels as both “crack” and “busbar,” the model was forced to choose only one. Figures \ref{fig:pv-vision_no_multi_label} and \ref{fig:pv-vision_no_multi_label_thresholding} illustrate this issue in detail.

In Figure \ref{fig:pv-vision_no_multi_label}, panel (a) shows the original EL image, while panel (b) presents the model’s predicted segmentation map, with each pixel assigned to one of five mutually exclusive classes. The middle row (c, d) contains the “double mask” targets we created for training: binary masks for crack and busbar features, where overlap pixels were labeled as both classes. The bottom row (e, f) shows the model’s raw activation maps for these features, where brighter regions correspond to higher predicted probabilities. Ideally, pixels at crack-busbar intersections would be bright in both maps, but here, high activation in one channel coincides with suppressed activation in the other. This demonstrates the “sigmoid bottleneck” effect \cite{grivasTamingSigmoidBottleneck2024}, where secondary class activations are weak and indistinguishable from background noise, even when using sigmoid outputs and explicit multi-label targets.

Figure \ref{fig:pv-vision_no_multi_label_thresholding} further quantifies this problem. The right panel shows the raw model output for a full image, with the dashed horizontal line at y = 130 marking the scanline used for the left plot. In that plot, each curve represents the activation values of a given class along the x-axis at y = 130. When the crack signal (pink) overlaps with the busbar signal (green), the busbar activation drops sharply, essentially disappearing into the noise floor and making multi-labeling infeasible.

This low secondary activation persisted through attempts of full model retraining, and training on doubled masks with intersection points labeled as both cracks and busbars. Finally, we were able to achieve co-classification of individual pixels with the multi-channel architecture described in Section \ref{sec:methods}, where each feature had its own dedicated input and output channel.

\subsection{Future Work}
\label{sec:future_work}

This work lays the groundwork for several impactful future directions. In addition to model extension to new EL feature classes, our multi-hot classification approach enables accurate prediction of power loss associated with quantified degradation image features. We lend our segmentation approach in service of predicting power loss due to cell cracks \cite{braidPowerLossEstimation2025, jostSolarCellCrack2024a}.

The approach employed in that work is to parameterize cell cracks, dark areas and busbars using a variational autoencoder (VAE). The resulting latent vector representation of the crack, dark and busbar masks is then used to predict power loss with ana XG-Boostregression model. This segmentation and parameterization approach gives better mechanistic context for a power loss model, preliminary results can be seen in references \cite{braidPowerLossEstimation2025, jostSolarCellCrack2024a}. The pvcracks repository also hosts examples \cite{jostPvcracks2024}. Furthermore, the VAE considers the latent vector as a standalone representation of the input/output image(s), so it can also be used to generate representative synthetic data. Owhereas other deep learning approaches, that we are working on, consider the entire EL module image \cite{byfordAdvancedPhotovoltaicModule2025}, future version of this model use electrical circuit simulations for the generation of synthetic module IV data for training. Furthermore, the VAE considers the latent vector as a standalone representation of the input/output image(s), so it can also be used to generate representative synthetic data.

Finally, future work could investigate specialized techniques to handle class imbalance. Potential approaches include: (1) focal loss to dynamically adjust the contribution of easy versus hard examples during training, (2) class-weighted loss functions where weights are inversely proportional to pixel-level class frequencies, (3) targeted augmentation strategies that oversample or synthetically generate additional examples of rare defect types, and (4) cost-sensitive learning frameworks that assign asymmetric penalties to false negatives versus false positives based on the operational impact of missing critical defects.

\section{Conclusion}
\label{sec:conclusion}

The results presented in this paper demonstrate a significant advancement in the automated segmentation and classification of EL images of PV cells. Most notably, our model achieves state-of-the-art accuracy even in complex scenarios where multiple features such as cracks and busbars overlap spatially. This is a critical for photovoltaic module degradation characterization and research, as it enables accurate quantification of the presence and interaction of multiple features which can affect power output of a PV cell.

A core strength of our approach lies in the adoption of a multi-channel U-Net architecture, which allows our model to independently activate multiple classes per pixel, thereby overcoming the mutual exclusivity limitation of softmax-based or sigmoid-based one-hot segmentation models. Our results confirm that sigmoid-based multi-hot segmentation on multiple channels, when trained on high-quality labeled datasets, can robustly co-label overlapping features such as cracks intersecting busbars. 

Moreover, our model generalizes across datasets collected from various institutions, despite differences in imaging formats and module types. This robustness speaks to the power of a learning-based approach over classical image processing methods, which are typically sensitive to such variations. 

Our multi-channel segmentation method can be employed to enhance O\&M strategies for large-scale photovoltaic installations, especially when combined with autonomous inspection methods such as electroluminescence imaging. Rapid statistical analysis of defects in PV modules will enable improved prediction of system health and lifetime, thus ensuring a more secure, independent, and safe energy supply.

\section{Acknowledgments}
Funding provided as part of the Durable Module Materials Consortium (DuraMAT), an Energy Materials Network Consortium funded by the U.S. Department of Energy, Office of Critical Minerals and Energy Innovation, Integrated Energy Systems Office under agreement number 32509. The views expressed in the article do not necessarily represent the views of the DOE or the U.S. Government. The U.S. Government retains and the publisher, by accepting the article for publication, acknowledges that the U.S. Government retains a nonexclusive, paid-up, irrevocable, worldwide license to publish or reproduce the published form of this work, or allow others to do so, for U.S. Government purposes. Sandia National Laboratories is a multimission laboratory managed and operated by National Technology and Engineering Solutions of Sandia, LLC, a wholly owned subsidiary of Honeywell International Inc., for the U.S. Department of Energy's National Nuclear Security Administration under contract DE-NA0003525.


\bibliographystyle{elsarticle-num}
\bibliography{references}

\end{document}